\documentclass[letterpaper, 10 pt, conference]{ieeeconf}  % Comment this line out if you need a4paper

\IEEEoverridecommandlockouts                              % This command is only needed if 
\usepackage{graphics} % for pdf, bitmapped graphics files
\usepackage{graphicx}
\usepackage{multirow}

\usepackage{subcaption}
\usepackage{makecell}
\usepackage{soul}
\usepackage{url}
\usepackage{booktabs}
\usepackage[table]{xcolor}
\soulregister\cite7
\soulregister\ref7

\title{\LARGE \bf
\name: Push High-Efficiency Optical Flow To the Limit
}

\newcommand{\name}{NeuFlow-V2}
\newcommand{\predecessor}{NeuFlow-V1}

\author{
Zhiyong Zhang$^{1*}$, Aniket Gupta$^{2*}$, Huaizu Jiang$^{2}$, Hanumant Singh$^{1}$%
\thanks{$^{1}$Department of Electrical and Computer Engineering, Northeastern University, Boston, MA 02115.}%
\thanks{$^{2}$Khoury College of Computer Sciences, Northeastern University, Boston, MA 02115.}%
\thanks{$^*$Equal contribution.}%
\thanks{Huaizu Jiang is supported by the National Science Foundation under Award IIS-2310254.}%
\thanks{\texttt{\{zhang.zhiyo, gupta.anik, h.jiang, ha.singh\}@northeastern.edu}}
}

\begin{document}

\maketitle
\thispagestyle{empty}
\pagestyle{empty}

%%%%%%%%%%%%%%%%%%%%%%%%%%%%%%%%%%%%%%%%%%%%%%%%%%%%%%%%%%%%%%%%%%%%%%%%%%%%%%%%
\begin{abstract}
% Real-time high-accuracy optical flow estimation is critical  for a number of real-world  robotic applications. While recent learning-based optical flow methods have achieved high accuracy, they often come with significant computational costs. In this paper, we propose a highly efficient optical flow method that balances high accuracy with reduced computational demands. Building upon NeuFlow v1, we introduce new components including a much more light-weight backbone and a fast refinement module. Both these modules help in keeping the computational demands light while providing close to state of the art accuracy. Compares to other state of the art methods, our model achieves a 10x-70x speedup while maintaining comparable performance on both synthetic and real-world data. It is capable of running at over 20 FPS on 512x384 resolution images on a Jetson Orin Nano. The full training and evaluation code is available at \url{https://github.com/neufieldrobotics/NeuFlow_v2}.

Real-time high-accuracy optical flow estimation is critical for a variety of real-world robotic applications. However, current learning-based methods often struggle to balance accuracy and computational efficiency: methods that achieve high accuracy typically demand substantial processing power, while faster approaches tend to sacrifice precision. These fast approaches specifically falter in their generalization capabilities and do not perform well across diverse real-world scenarios. 
In this work, we revisit the limitations of the SOTA methods and present \name, a novel method that offers both — high accuracy in real-world datasets coupled with low computational overhead. 
In particular, we introduce a novel light-weight backbone and a fast refinement module to keep computational demands tractable while delivering accurate optical flow. 
Experimental results on synthetic and real-world datasets demonstrate that \name\ provides similar accuracy to SOTA methods while achieving 10x-70x speedups.
It is capable of running at over 20 FPS on 512x384 resolution images on a Jetson Orin Nano. The full training and evaluation code is available at \url{https://github.com/neufieldrobotics/NeuFlow_v2}.

% Real-time high-accuracy optical flow estimation is crucial for various real-world applications. While recent learning-based optical flow methods have achieved high accuracy, they often come with significant computational costs. In this paper, we propose a highly efficient optical flow architecture that addresses both high accuracy and computational cost concerns. The model builds upon NeuFlow v1, incorporating new components such as an extremely lightweight backbone and a simple RNN module that improve refinement accuracy while reducing computational demands. We evaluate our approach on the Jetson Orin Nano and RTX 2080 \aniket{Why do we need evaluation on RTX 2080?}, demonstrating significant efficiency improvements \hl{across multiple GPU platforms}. Our model achieves a 10×-70× speedup compared to several state-of-the-art methods while maintaining comparable accuracy on the KITTI and Sintel datasets. Our approach reaches approximately 10 FPS on edge computing platforms, marking a significant breakthrough in deploying complex computer vision tasks such as SLAM on small robots. \aniket{This line seems too much. Writing something like, with improved model architecture, the model generalizes much better on real-world data is sufficient} The full training and evaluation code is available at https://github.com/neufieldrobotics/NeuFlow\_v2.

\end{abstract}

%%%%%%%%%%%%%%%%%%%%%%%%%%%%%%%%%%%%%%%%%%%%%%%%%%%%%%%%%%%%%%%%%%%%%%%%%%%%%%%%

\section{Introduction}
Optical flow is the apparent motion of objects, surfaces, or edges in a visual scene caused by the relative movement between the camera and the scene. It represents a dense field of motion vectors that describe how each pixel in an image moves between two consecutive frames. This problem remains unsolved for edge devices which are particularly employed on fast-moving robots like drones. The best algorithms are limited by their generalization capabilities across diverse scenarios, fast-moving objects, occlusions etc. 

Optical flow algorithms have seen substantial progress in recent years \cite{zhai2021optical}. Starting from FlowNet \cite{dosovitskiy2015flownet}, learning-based methods for optical flow have shifted towards feature learning for matching, moving away from traditional hand-crafted features like those in Lucas-Kanade \cite{lucas1981iterative} or SIFT \cite{lowe2004distinctive} \cite{liu2008sift}. Despite these advances, early optical flow methods that rely on CNNs struggle with significant challenges such as handling large displacements \cite{sun2018pwc}. Furthermore, their one-shot architectures do not generalize well to real-world data \cite{teed2020raft} \cite{dosovitskiy2015flownet}, \cite{sun2018pwc},  \cite{ilg2017flownet}, \cite{ranjan2017optical}. %early optical flow methods struggled with significant challenges such as large displacement and generalizing to real-world data \cite{teed2020raft}. 
This can be primarily attributed to the difficulty in collecting ground truth optical flow data in the real-world, simulation is predominantly used to generate sufficient training data \cite{dosovitskiy2015flownet}, \cite{mayer2016large}. However, training with simulation data can lead to overfitting due to unrealistic illumination, reflections, and monotonous scenes \cite{wang2020tartanair}, \cite{mehl2023spring}. 
 
% Some recent deep learning approaches have mitigated these issues, .
% \hj{Let's merge the first two paragraphs and make a quick introduction of learning-based approaches.} \aniket{Done}

Starting with RAFT \cite{teed2020raft}, iterative refinements have partially mitigated the generalization issue while also capturing larger motions \cite{xu2022gmflow}, \cite{jiang2021learning} albeit with an increased computational cost \cite{sui2022craft}, \cite{huang2022flowformer}. Recent research has further improved accuracy and generalization by incorporating the latest modules, such as Transformer \cite{zhao2022global}, Partial Kernel Convolution \cite{morimitsu2024recurrent}, Super Kernel \cite{sun2022skflow} etc. However, these  methods are generally more computation heavy due to the iterative refinement process. Some models need over 30 iterations to generate a stable optical flow \cite{teed2020raft}, while others reduce the number of iterations but increase the computational load of each iteration \cite{morimitsu2024recurrent}, \cite{wang2024sea}.
% \hj{How about the computation burden of the backbone?}

\begin{figure}[t]
\hfill
\begin{center}
\includegraphics[width=\linewidth]{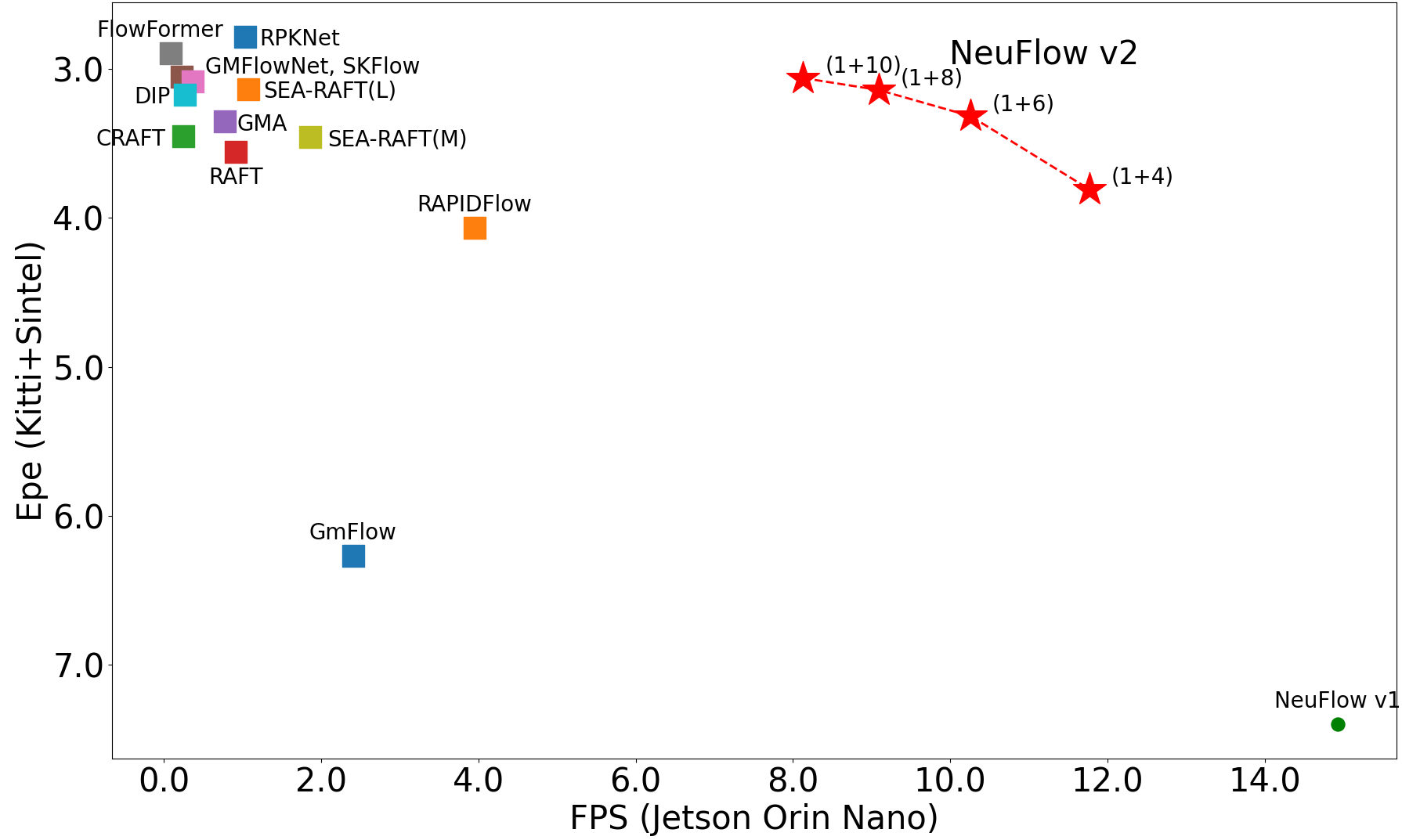}
\end{center}
\caption{End Point Error (EPE) of KITTI and Sintel datasets vs. Frames Per Second (FPS) throughput on an edge computing platform (Jetson Orin Nano). Individual points represent a broad class of optical flow methods. Our algorithm is comparable in accuracy but significantly more efficient, approaching an order of magnitude improvement in computational complexity. All models were trained solely on the FlyingThings datasets. For the notation ``\name\ $(x+y)$'', $x$ represents number of refinement iterations at 1/16th scale and $y$ represents number of refinement iterations at 1/8th scale.
% \hj{Use a different legend (not star) for NeuFlow-V1 to make it different from v2.}\aniket{Addressed} 
% \aniket{The RAPIDFlow paper also present it this way - so I think showing the iterations might be okay}
}
\vspace{-2.5em}
\label{neuflow_plot}
\end{figure}

\begin{figure*}[ht]
\begin{center}
\includegraphics[width=\textwidth]{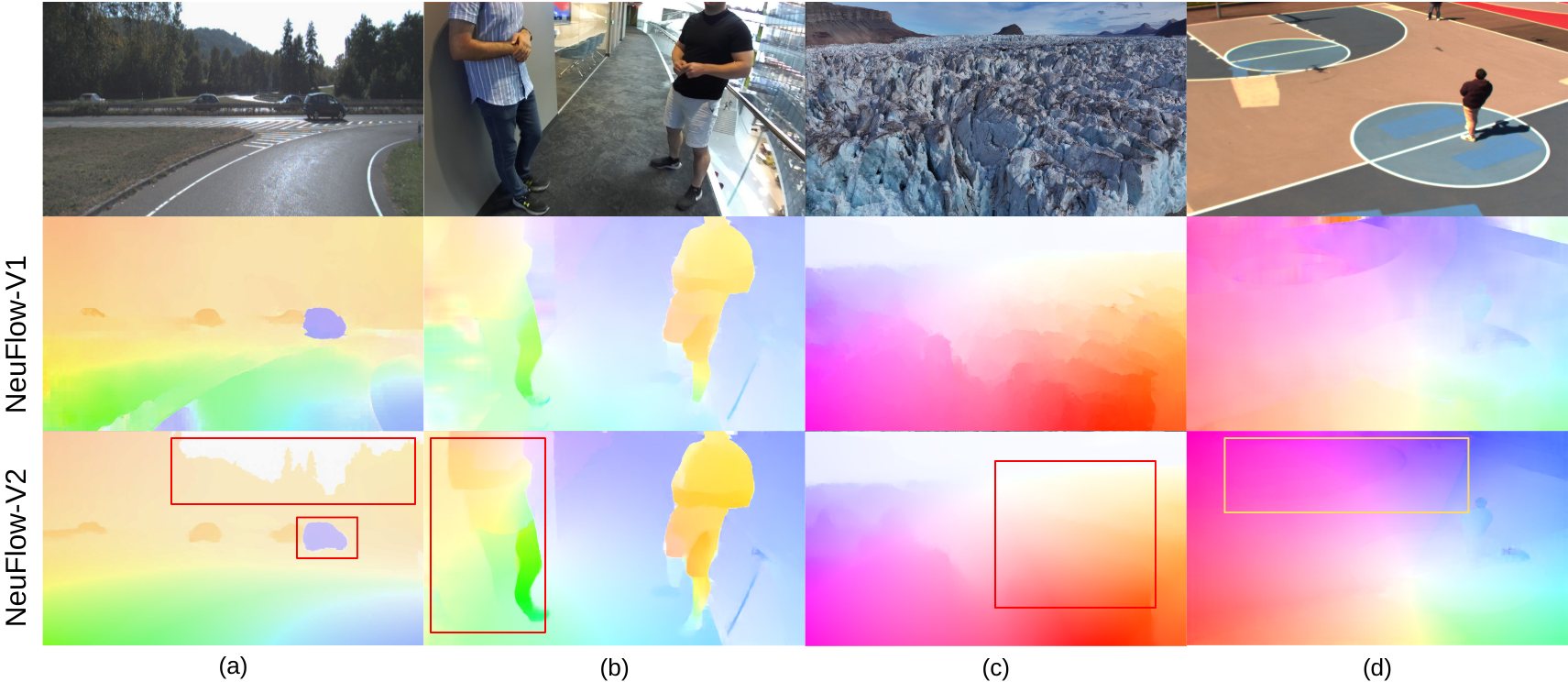}
\end{center}
\caption{Optical flow results on unseen real-world images. 
% : We run \name on unseen real-world images to showcase the model's generalization capabilities. 
In comparison to \predecessor, \name\ outputs more accurate optical flow at capturing the scene details (highlighted with bounding boxes in (a) and (b)) and also much smoother and more accurate results in (c) and (d). 
% \name\ is also more accurate at capturing the scene details better - see (a) and (b).
% \hj{Add another row to show results of NeuFlow v1 so readers can get a better sense of the improvement.}
% \zz{I don't think we should put too much emphasize on the difference between version 1 and version 2, because there is some generalization problem with version 1}
% \hj{But the generalization issue of V1 is what we are addressing here in V2?}
% \hj{Can we add circles to highlight places where V2 works better. For example, tops of trees in the first column and around the first person in the second column?}\aniket{Addressed}
}
\vspace{-1.5em}
\label{neuflow_examples}
\end{figure*}

\predecessor~\cite{zhang2024neuflow} provides real-time inference speeds on edge devices like Jetson Orin Nano for optical flow estimation. However, it faces notable challenges in generalizing to real-world datasets. Although incorporating refinement modules from prior works \cite{sui2022craft, xu2022gmflow} in the \predecessor\ architecture could address these issues, they significantly increase computational overhead, limiting real-time feasibility. To overcome this, we present an efficient iterative refinement module that leverages only CNN layers arranged in a recurrent fashion. This helps mitigate the generalization issue while maintaining fast inference speed on edge devices. To further optimize the \predecessor\ architecture, we introduce a new backbone that fuses multi-scale features more effectively, achieving a ~24\% (backbone inference time)
% \hj{What is pp? And why only measure backbone time?}\aniket{pp is percentage points, Just to show that the current backbone is faster and more suitable for the job. I can add the total time as well if that makes more sense} 
speedup compared to the original design while improving accuracy. 
% \hj{It contains little information of how NeuFlow-v2 is different from v1. Thus it is hard to get the significance of the contributions. Explain why the modifications of the backbone and refinement module are important and how they are different from v1.} \aniket{So technically the only two differences are the faster backbone and the iterative refinement}

We evaluate \name\ extensively on both synthetic \cite{butler2012naturalistic, mayer2016large} and real-world \cite{geiger2012we} datasets. \name\ achieves real-time inference while delivering near state-of-the-art accuracy on edge devices, making it well-suited for deployment on small robotic platforms. Fig. \ref{neuflow_plot} plots the end point error (EPE) vs FPS for the latest optical flow methods. \name\ significantly outperforms \predecessor\ while still maintaining real-time speeds. Fig. \ref{neuflow_examples} shows generalization examples of \name\ on unseen real-world data in comparison with \predecessor. 

In summary, we make the following two critical contributions in devising \name.
\begin{itemize}
    \item Lightweight Backbone: A simple CNN-based backbone for extracting low-level features from multi-scale images. Unlike commonly used architectures such as ResNet \cite{he2016deep} or Feature Pyramid Networks \cite{lin2017feature}, this efficient backbone is sufficient for accurate optical flow estimation.

    \item Efficient Iterative Refinement Module: A lightweight recurrent module that refines optical flow predictions while maintaining efficiency. Instead of computationally expensive approaches like LSTMs \cite{hochreiter1997long} or GRUs \cite{cho2014learning}, our simplified refinement achieves higher accuracy with minimal overhead.
\end{itemize}

\section{Related Work}

FlowNet \cite{dosovitskiy2015flownet} was the first deep learning-based optical flow estimation method, introducing two variants: FlowNetS and FlowNetC, along with the synthetic FlyingChairs dataset for end-to-end training and benchmarking. An improved version, FlowNet 2.0 \cite{ilg2017flownet}, fused cascaded FlowNets with a small displacement module, decreasing the estimation error by more than 50\% while being marginally slower.

Following FlowNet 2.0 \cite{ilg2017flownet}, researchers developed more lightweight optical flow methods. SPyNet \cite{ranjan2017optical} is 96\% smaller than FlowNet in terms of model parameters. PWC-Net \cite{sun2018pwc} is 17 times smaller than FlowNet 2. LiteFlowNet \cite{hui2018liteflownet} is 30 times smaller in model size and 1.36 times faster in running speed compared to FlowNet 2. LiteFlowNet 2 \cite{hui2020lightweight} improved optical flow accuracy on each dataset by around 20\% while being 2.2 times faster. LiteFlowNet 3 \cite{hui2020liteflownet3} further enhanced flow accuracy. RapidFlow \cite{morimitsu2024rapidflow} combines efficient NeXt1D convolution blocks with a fully recurrent structure to decrease computational costs. DCVNet \cite{jiang2023dcvnet} proposes constructing cost volumes with different dilation factors to capture small and large displacements simultaneously. \predecessor \cite{zhang2024neuflow}, our previous work, is a fastest optical flow method, being over ten times faster than mainstream optical flow methods while maintaining comparable accuracy on the Sintel and FlyingThings datasets.

\begin{figure*}[t]
\begin{center}
\includegraphics[width=\textwidth]{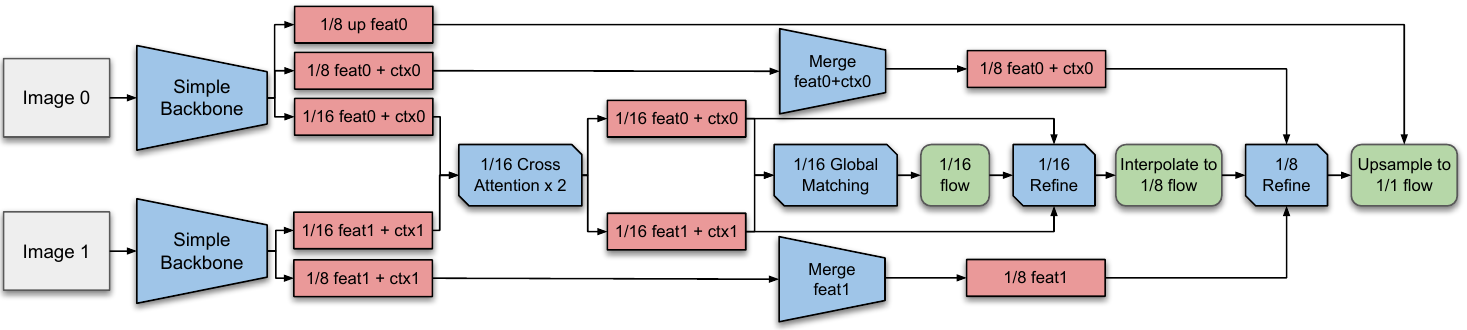}
\end{center}
 \caption{\name\ architecture.
 % , which estimates a coarse optical flow using global attention followed by iterative local refinement. 
 We begin with a simple CNN backbone that outputs features and context at 1/8 and 1/16 scales for both images. 
 The features at the 1/16 scale are then fed into cross-attention layers for feature enhancement. Next, we perform global matching to obtain an initial flow at the 1/16 scale, which is refined through one iteration. This flow is upsampled to a 1/8 scale and further refined over eight iterations. The refined 1/8-scale flow is then upsampled to full resolution using a convex upsampling module. }
\vspace{-1em}
\label{neuflow_arct}
\end{figure*}

\begin{figure*}[ht]
\begin{center}
\includegraphics[width=\textwidth]{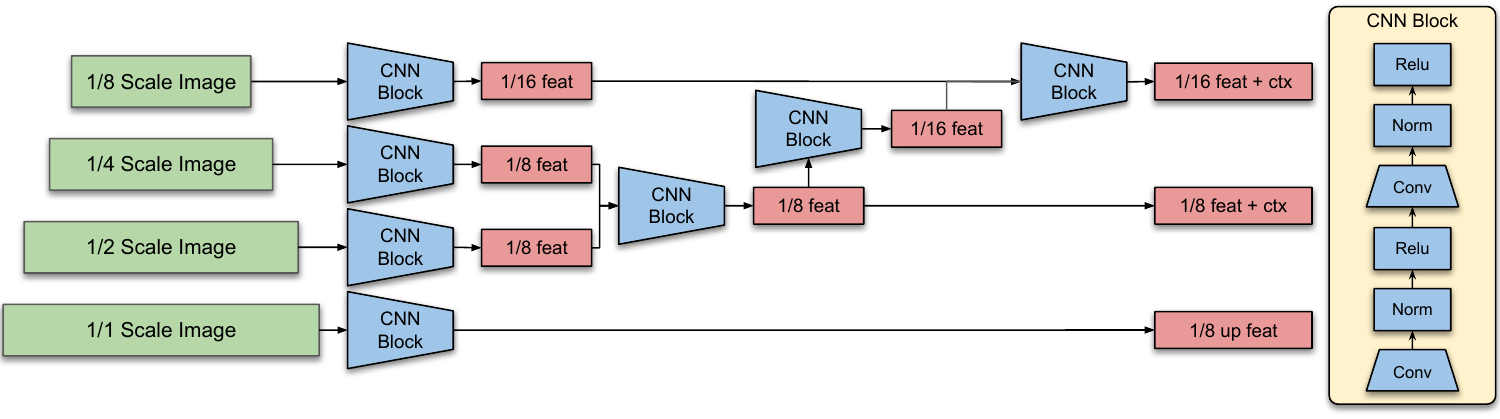}
\end{center}
\caption{\name\ simple backbone. We downsample the image into various scales, ranging from 1/1 to 1/8. A simple CNN block, consisting of two [Convolution, Norm, ReLU] layers, is used to extract low-level features from the image at various scales. Then, using the same CNN block design, we merge and resize these features into 1/8 and 1/16 scale outputs. The backbone outputs 1/8 and 1/16-scale features and context for further flow estimation, along with an additional 1/8-scale feature for convex upsampling.
% \hj{It can be significantly shrinked to save some space.}
}
\vspace{-1.5em}
\label{neuflow_backbone}
\end{figure*}

More recently, RAFT \cite{teed2020raft} used recurrent all-pairs field transforms to achieve strong cross-dataset generalization. Following RAFT, GMA \cite{jiang2021learning} introduced a global motion aggregation module to improve estimation in occluded regions. GMFlow \cite{xu2022gmflow} reformulated optical flow as a global matching problem. GMFlowNet \cite{zhao2022global} efficiently performed global matching by applying argmax on 4D cost volumes. CRAFT \cite{sui2022craft} used a Semantic Smoothing Transformer layer to make features more global and semantically stable. FlowFormer \cite{huang2022flowformer}, \cite{shi2023flowformer++} encodes the 4D cost tokens into a cost memory with alternate-group transformer layers in a latent space. SKFlow \cite{sun2022skflow} benefits from super kernels to complement the absent matching information and recover the occluded motions. DIP \cite{zheng2022dip} introduced the first end-to-end PatchMatch-based method, achieving high-precision results with lower memory. RPKNet \cite{morimitsu2024recurrent} utilized Partial Kernel Convolution layers to produce variable multi-scale features and efficient Separable Large Kernels to capture large context information. Sea-Raft \cite{wang2024sea} proposed a new loss (mixture of Laplace) and directly regressed an initial flow for faster convergence. Many works have also been proposed to either reduce computational costs or improve flow accuracy \cite{kong2021fastflownet}.

\section{Proposed Approach: NeuFlow v2}

% \hj{I suggest to add a section of ``Overview'', which explains the overall pipeline of NeuFlow v2 and its similarities and differences from NeuFlow v1. And talk about each major difference separately in the rest of this section.}
% We introduce NeuFlow v2, an improved version that maintains the highest inference speed while approaching state-of-the-art accuracy across multiple datasets. We developed a new simple backbone module that efficiently extracts low-level features for optical flow tasks, and a simple RNN refinement module that uses only CNNs, without GRU or LSTM, improving both accuracy and efficiency. With additional efficient modules, including cross-attention and global matching, we have created a stable optical flow model capable of running in real-time on edge devices.
We introduce \name, an enhanced version of \predecessor\ that preserves real-time inference speed on edge devices while achieving near state-of-the-art accuracy on real-world datasets. To accomplish this, we propose two key innovations: (1) a fast and efficient backbone that optimally balances inference speed and accuracy, and (2) a lightweight iterative refinement module that significantly improves \name's generalization to real-world datasets. The overall architecture of our model is illustrated in Fig. \ref{neuflow_arct}.
% \hj{Use Fig.~\ref{neuflow_arct} to give a brief overview.} \aniket{Addressed}

\subsection{Backbone}
% \hj{Use Fig.~\ref{neuflow_backbone} to explain the backbone. Similarly, please add references to other figures in the text.} \aniket{Addressed}
As presented in \cite{zhang2024neuflow}, a shallow backbone extracting low-level features is sufficient for computing pixel similarity and computing optical flow. We build on this insight and propose our simple backbone architecture. As presented in Fig.~\ref{neuflow_backbone} our backbone only extracts features from 1/2th, 1/4th and 1/8th images using a CNN block composed of convolution, normalization, and ReLU layers. This same CNN block is used to concatenate and resize these features into the desired output scale, specifically to 1/16-scale features and context, as well as 1/8-scale features and context. Features are used for correlation computation, while context is used for flow refinement.

Note that the 1/1-scale image is used solely for convex upsampling. In our experiments, we find that extracting features from 1/1-scale images leads to overfitting on the training data (FlyingThings) and thus are ineffective in flow computation on unseen data (Sintel, KITTI). See \ref{sec: expt/ablation} for details.

\subsection{Cross-Attention And Global Matching}

Cross-attention is used to exchange information between images globally, enhancing the distinctiveness of matching features and reducing the similarity of unmatched features. Global matching is then applied to find corresponding features globally, enabling the model to handle large pixel displacements, such as in fast-moving camera situations. To reduce the computational burden of cross-attention and global matching, we operate on 1/16-scale features instead of the 1/8-scale.

Similar to \predecessor\ and GMFlow+ \cite{xu2023unifying}, we utilize Transformers to implement cross-attention and global matching. While flow self-attention is typically used to refine pixels based on self-similarity after global matching \cite{xu2023unifying, xu2022gmflow}, we instead use our refinement module to iteratively refine the estimated flow.

\begin{figure}[t]
\begin{center}
\includegraphics[width=\linewidth]{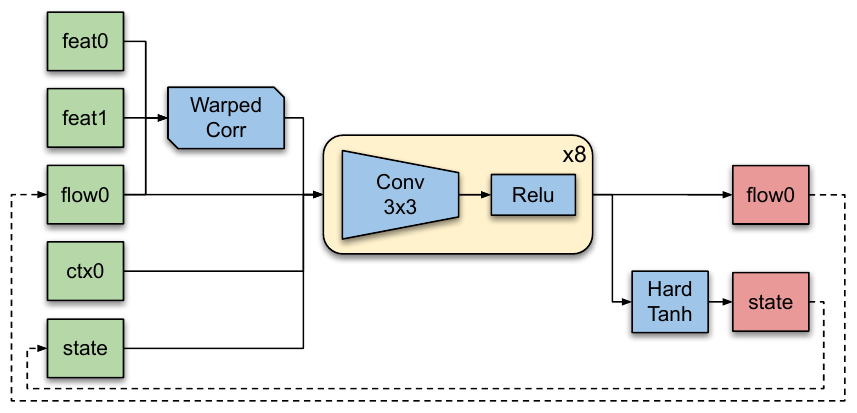}
\end{center}
\caption{\name\ iterative refinement. We first compute the correlation within nearby pixels and warp these values using the currently estimated flow. The warped correlation, current estimated flow, context features, and hidden state are then fed into a series of $3\times 3$ convolution layers followed by ReLU activation, repeated eight times. At the end of these layers, the network outputs both the refined flow and an updated hidden state for the next iteration. Instead of using GRU or LSTM modules, we use a simpler recurrent design without any gates. The \texttt{HardTanh} function is applied to update the hidden state, which mitigates the saturation issue related to the \texttt{tanh} function.
}
% \vspace{-2em}
\label{neuflow_refine}
\end{figure}

\subsection{Iterative Refinement Module}

We first compute the correlation within nearby $9\times9$ neighborhoods and warp this correlation using the estimated flow. We then concatenate the warped correlation, context features, estimated flow, and previous hidden state, processing them through eight layers of $3\times3$ convolutional layers followed by ReLU activation to output the refined optical flow and updated hidden state. Refinement is performed for the flow at both the 1/16 and 1/8 scales. Fig. \ref{neuflow_refine} illustrates the architecture of our refinement module.

To address the vanishing or exploding gradient problem, most RNNs use GRU or LSTM modules to compute the current hidden state based on the previous hidden state and current inputs (warped correlation, context, and flow) and to decode the estimated flow using the current hidden state. 
However, GRU and LSTM may be computationally heavy in a real-time method. 
In our design, we adopt a simpler design without any gates in a recurrent model, where we only have a set of $3\times 3$ convolution layers followed by ReLU.

% modules often have too few layers to effectively merge the current input with the previous hidden state.
% \hj{Contradicts with the previous statement that GRU and LSTM are computationally heavy.}
% In contrast, we use deep CNN layers to effectively merge the inputs (warped correlation, context, and flow) with the hidden state. This approach has been tested to avoid unstable gradient issues and significantly improve accuracy, as detailed in the ablation study in Section \ref{sec: expt/ablation}.

The \texttt{tanh} function is commonly applied to the output hidden state, keeping values within the range (-1,1). However, when hidden state values approach these limits, the corresponding inputs to \texttt{tanh} in the previous layer can become extremely large or small, leading to numerical overflow. 
%To mitigate this, we use \texttt{HardTanh}~\cite{collobert2011natural} to constrain feature values within a specific range and improve numerical stability. By clipping extreme values to [-1,1], \texttt{HardTanh} prevents large exponentials and mitigates overflow issues.
To mitigate the gradient vanishing/explosion issue related to the saturation of the \texttt{tanh} activation function, we use the \texttt{HardTanh}~\cite{collobert2011natural} for the hidden state update.
We clip the input to be in the range of [-4, 4], where there are reasonable gradients values.
As shown in the ablation study in Section \ref{sec: expt/ablation}, our simple design does not only run faster but also produces more accurate optical flow than using GRU.

% \hj{Please confirm this section.}

\begin{figure}[t]
\begin{center}
\includegraphics[width=\linewidth]{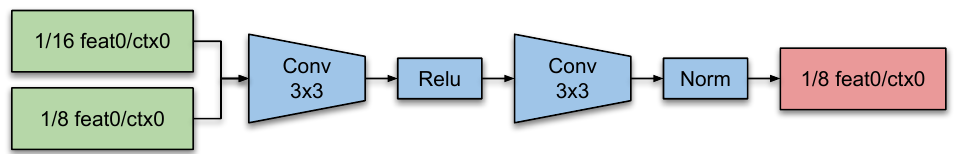}
\end{center}
\caption{\name\ merge module. We concatenate the 1/8-scale features/context with the interpolated 1/16-scale features/context and use simple CNN blocks to output 1/8-scale features/context, thereby incorporating both global and local information.
% \hj{No norm after the first conv and no ReLu after the second Conv+Norm?} \aniket{That's what we have in the model - ZZ can you explain the design choice here?}
}
\label{neuflow_merge}
\end{figure}

\subsection{Multi-Scale Feature/Context Merge}

Our simple backbone lacks the depth of convolution, resulting in features having a small receptive field. Cross-attention at the 1/16-scale provides global attention, offering a global feature/context. However, 1/8-scale features/context do not have a global receptive field. Therefore, we merge the 1/16-scale global features/context with the 1/8-scale local features/context to ensure that the 1/8-scale features/context contain both global and local information.

The merge block (Fig. \ref{neuflow_merge}) consists of two layers of CNNs with ReLU activation and normalization. In practice, features and context are merged individually using the same merge block structure.

\section{Experiments}

\begin{table*}[ht]
\centering
\begin{tabular}{cl|ccccc|ccccc}
\toprule
 & \multirow{2}{*}{\multirow{2}{*}{Method}} & \multicolumn{2}{c}{Sintel (train)} & \multirow{2}{*}{\multirow{2}{*}{\makecell{RTX\\2080 (s)}}} & \multirow{2}{*}{\multirow{2}{*}{\makecell{Jetson Orin\\Nano (s)}}} & \multirow{2}{*}{\multirow{2}{*}{\makecell{Batch\\Size (8G)}}} & \multicolumn{2}{c}{KITTI-15} & \multirow{2}{*}{\multirow{2}{*}{\makecell{RTX\\2080 (s)}}} & \multirow{2}{*}{\multirow{2}{*}{\makecell{Jetson Orin\\Nano (s)}}} & \multirow{2}{*}{\multirow{2}{*}{\makecell{Batch\\Size (8G)}}} \\
\cmidrule(r){3-4} \cmidrule(r){8-9}
 &  & clean & final &  & & & EPE & F1 & & \\
\midrule
\multirow{8}{*}{\rotatebox{90}{\textbf{Slow}}}
    & SKFlow (32 iters) & 1.22 & 2.46 & 0.365 & 4.181 & 17 & 4.27 & 15.5 & 0.408 & 4.478 & 16 \\
    & SEA-RAFT(M) (4 iters) & 1.21 & 4.04 & \textbf{0.061} & \textbf{0.524} & 23 & 4.29 & 14.2 & \textbf{0.075} & \textbf{0.549} & 22 \\
    & DIP (20 iters) & 1.30 & 2.82 & 0.499 & 3.615 & 12 & 4.29 & 13.7 & 0.523 & 3.767 & 12 \\
    & GMFlowNet (32 iters) & 1.14 & 2.71 & 0.227 & 2.626 & 12 & 4.24 & 15.4 & 0.231 & 2.761 & 12 \\
    & FlowFormer & \textbf{1.01} & \textbf{2.40} & 1.007 & 11.196 & 4 & 4.09 & 14.7 & 1.002 & 11.172 & 5 \\
    & RPKNet (12 iters) & \underline{1.12} & \underline{2.45} & 0.158 & 0.947 & \textbf{68} & \underline{3.79} & \underline{13.0} & 0.157 & 0.976 & \textbf{64} \\
    & SEA-RAFT(L) (12 iters) & 1.19 & 4.11 & \underline{0.096} & \underline{0.910} & 22 & \textbf{3.62} & \textbf{12.9} & \underline{0.101} & \underline{0.953} & 20 \\
    & GMA (32 iters) & 1.30 & 2.74 & 0.152 & 1.245 & 17 & 4.69 & 17.1 & 0.161 & 1.343 & 16 \\
    & RAFT (32 iters) & 1.43 & 2.71 & 0.125 & 1.060 & \underline{24} & 5.04 & 17.4 & 0.129 & 1.126 & \underline{23} \\
    & CRAFT (32 iters) & 1.27 & 2.79 & 0.347 & N/A & 2 & 4.88 & 17.5 & 0.385 & N/A & 2 \\
\midrule
\multirow{5}{*}{\rotatebox{90}{\textbf{Fast}}} 
    & RAPIDFlow (12 iters) & 1.58 & \underline{2.94} & 0.038 & 0.252 & \underline{103} & \underline{5.87} & \underline{17.7} & 0.045 & 0.254 & \underline{99} \\
    & GMFlow (1 iter) & \underline{1.50} & 2.96 & 0.046 & 0.404 & 24 & 10.3 & 33.6 & 0.055 & 0.426 & 22 \\
    & \predecessor & 1.66 & 3.13 & \textbf{0.009} & \textbf{0.064} & \textbf{107} & 12.4 & 32.5 & \textbf{0.010} & \textbf{0.070} & \textbf{115} \\
    & \name (9 iters) & \textbf{1.24} & \textbf{2.67} & \underline{0.015} & \underline{0.106} & 26 & \textbf{4.33} & \textbf{15.3} & \underline{0.015} & \underline{0.114} & 21 \\
\bottomrule

\end{tabular}

\caption{This table compares the latest optical flow methods based on their highest accuracy. All models were trained on the FlyingThings dataset and evaluated on the Sintel and KITTI training sets. Inference time was measured on both an RTX 2080 and an edge computing device, the Jetson Orin Nano, using half-precision models. Batch size was determined during batch inference and both RTX 3080 and Jetson Nano used have 8GB memory. CRAFT is marked as N/A because the Jetson Orin Nano crashed during inference. The table shows that \name\ achieves the best accuracy in the Fast methods while still running in real-time. This table also highlights that \predecessor\ while being really fast, does not work well on real-world KITTI data. *Best performance is marked in bold and second best is marked with underline}
\label{table_1}
\end{table*}

\begin{figure}[t]
\hfill
\begin{center}
\includegraphics[width=\linewidth]{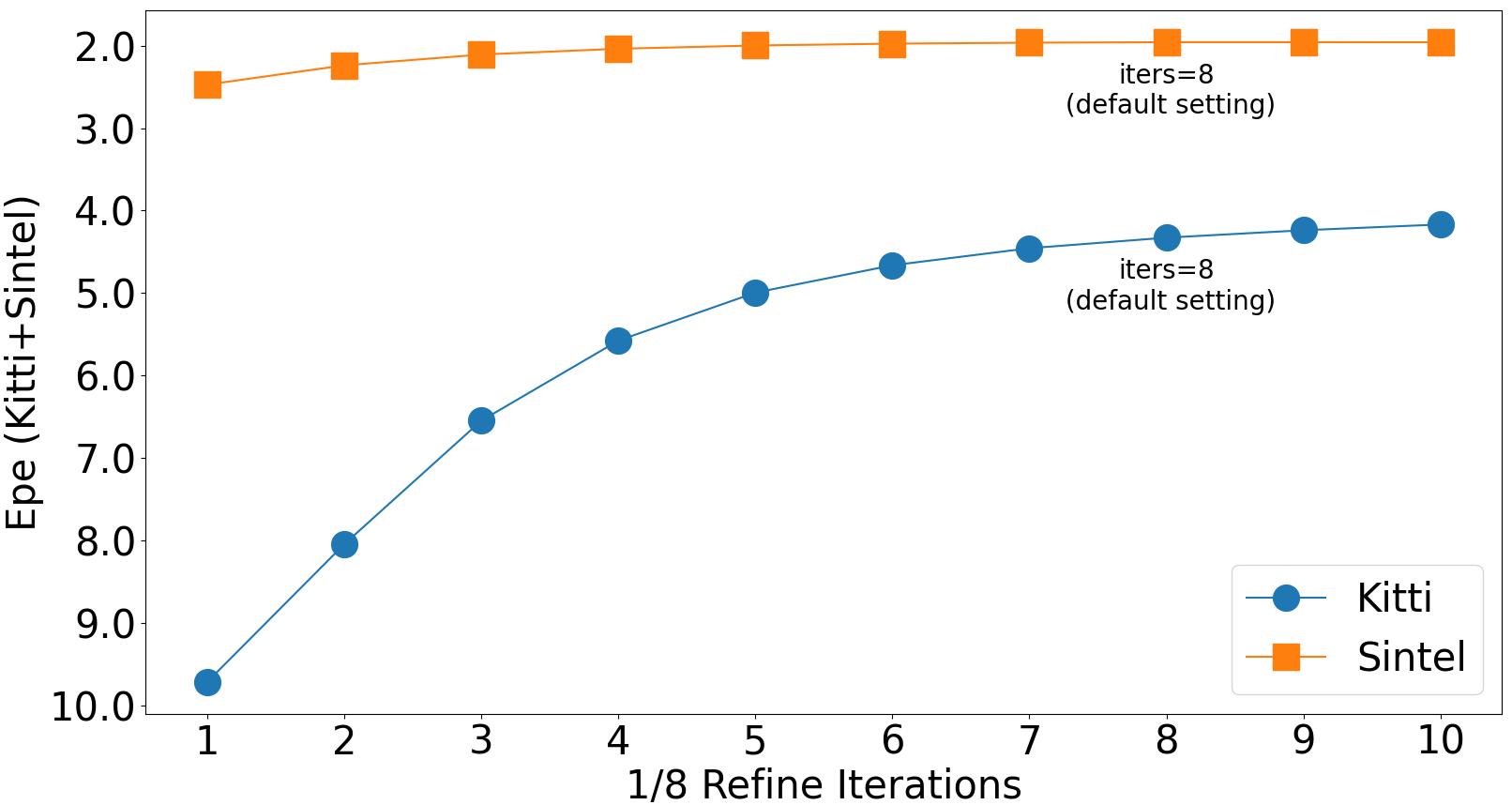}
\end{center}
\caption{End Point Error (EPE) for the KITTI and Sintel datasets vs. the number of 1/8 refinement iterations: The default configuration is 8 iterations. Accuracy on Sintel converges quickly within the first few iterations, while KITTI continues to improve until 8-10 iterations.}
\label{iters_plot}
\end{figure}

% \hj{There should be a table showing that if we start from NeuFlow v1, how much improvement we can get for each modification until we get NeuFlow v2.} \aniket{This is quite difficult to do right now with the resources and time we have. }

\subsection{Training and Evaluation Datasets}
We first train the model solely on the FlyingThings \cite{mayer2016large} dataset for a fair comparison with other models, as most models have undergone the same procedure.
% Additionally, we trained the model using a mixed dataset comprising Sintel \cite{butler2012naturalistic}, KITTI \cite{geiger2012we}, and HD1K for real-world applications.
For evaluation, we followed the common practice of using the Sintel and KITTI datasets to demonstrate the model's generalization capabilities. We followed the same procedure and data augmentation settings as RAFT, utilizing RAFT's \cite{teed2020raft} training and evaluation code.

% For real-world use cases, we also train the model with a mixed dataset that includes FlyingThings, Sintel, KITTI, HD1K, and VIPER \cite{richter2017playing}. The pretrained model is available in our repository. \aniket{Which model do we use in our evaluation? Why do we need to mention these three different models here?}

\subsection{Comparison with Latest Optical Flow Methods}

We compare our method against several state-of-the-art optical flow methods, including methods optimized for highest accuracy (Slow) and methods optimized for highest speed (Fast) in Tab. \ref{table_1}. To show the effectiveness of our approach across GPU platforms, we measure the computation time on Nvidia RTX 2080 GPU and the edge computing device Nvidia Jetson Orin Nano (8GB). We evaluate on the Sintel dataset with $1024 \times 436$ resolution and KITTI-15 dataset with $1242 \times 375$ resolution. The inference batch size is measured on the Jetson Orin Nano with 8GB memory to assess memory usage of all methods. 

Tab. \ref{table_1} shows that amongst all the methods \name\ achieves comparable performance to SEA-RAFT(M) \cite{wang2024sea} and SKFlow \cite{sun2022skflow} while being 5x-40x faster on the Sintel Dataset. On the KITTI dataset, \name\ achieves comparable performance to SKFlow\cite{sun2022skflow}, SEA-RAFT(M)\cite{wang2024sea}, DIP\cite{zheng2022dip} and FlowFormer\cite{huang2022flowformer} while being 5x-110x faster. Amongst fast methods, \name\ achieves the best accuracy on both datasets while being 2.5x faster than RAPIDFlow\cite{morimitsu2024rapidflow} and 4x faster than GMFlow\cite{xu2022gmflow}. In comparison to \predecessor, we achieve a huge jump in accuracy on the real-world KITTI dataset and significant improvements on the Sintel dataset - which justifies the choices we have made in designing \name\ over \predecessor.

\begin{table*}[ht]
\begin{center}
\begin{tabular}{lcc|ccc|ccc}
\toprule
\multirow{2}{*}{\multirow{2}{*}{Method}}& \multicolumn{2}{c}{Things} & \multicolumn{2}{c}{Sintel (train)} & \multirow{2}{*}{\multirow{2}{*}{\makecell{Jetson Orin\\Nano (s)}}} & \multicolumn{2}{c}{KITTI-15} & \multirow{2}{*}{\multirow{2}{*}{\makecell{Jetson Orin\\Nano (s)}}} \\
\cmidrule(r){2-3} \cmidrule(r){4-5} \cmidrule(r){7-8}
& train & val & clean & final & & EPE & F1 & \\
\midrule
Full & 2.97 & 3.44 & 1.24 & 2.67 & 0.106 & 4.33 & 15.3 & 0.114 \\
\midrule
\multicolumn{9}{c}{\textbf{Backbone Module}} \\
\midrule
YOLO v8 Backbone & 2.56 & 3.15 & 1.34 & 2.96 & 0.111 & 4.87 & 17.0 & 0.121 \\
1/1 backbone & 2.73 & 3.28 & 1.19 & 2.81 & 0.112 & 4.92 & 15.7 & 0.122 \\
\predecessor\ backbone & 2.77 & 3.34 & 1.24 & 2.87 & 0.116 & 4.98 & 15.8 & 0.128 \\
\midrule
\multicolumn{9}{c}{\textbf{Refine Module}} \\
\midrule
-2 layers & 2.85 & 3.41 & 1.22 & 2.76 & 0.099 & 5.21 & 16.5 & 0.106 \\
+2 layers & 2.67 & 3.17 & 1.18 & 2.80 & 0.113 & 4.62 & 15.6 & 0.122 \\
half feature dimension & 3.26 & 3.79 & 1.37 & 2.91 & 0.095 & 6.66 & 21.7 & 0.102 \\
use ConvGRU & 3.00 & 3.62 & 1.28 & 3.87 & 0.121 & 7.36 & 20.4 & 0.129 \\
\midrule
\multicolumn{9}{c}{\textbf{Architecture}} \\
\midrule
w/o cross attention & 3.99 & 4.37 & 1.60 & 4.17 & 0.098 & 5.18 & 16.3 & 0.105 \\
w/o global match, 1/16 refine=4 & 2.85 & 3.24 & 1.21 & 2.88 & 0.111 & 4.89 & 16.1 & 0.120 \\
w/o 1/16 refine & 3.00 & 3.79 & 1.40 & 4.06 & 0.103 & 5.52 & 19.6 & 0.111 \\
\bottomrule
\end{tabular}
\end{center}
\caption{Ablation Study: Starting with ablations on the backbone, we remove the full scale features from the backbone used in \predecessor\ as it leads to slightly better performance. We also experiment with adding/reducing the number of layers in the Iterative refinement module and find out that setting 8 refinement layers provides a good balance in accuracy and computation time. Removing Cross-attention or global matching as expected leads to drop in accuracy. We also find out that removing refinement on 1/16 scale also leads to a significant drop in accuracy}
\label{ablation}
\end{table*}

% \begin{table*}[ht]
% \begin{center}
% \begin{tabular}{|c|c|c||c|c|c||c|c|c|}
% \hline
% Refinement iterations & things train & things val & sintel clean & sintel final & \makecell{Jetson Orin\\ Nano (s)} & kitti epe & kitti F1 & \makecell{Jetson Orin\\ Nano (s)} \\
% \hline
% \textbf{1/16 iters=1, 1/8 iters=8} & \textbf{2.97} & \textbf{3.44} & \textbf{1.24} & \textbf{2.67} & \textbf{0.106} & \textbf{4.33} & \textbf{15.3} & \textbf{0.114} \\
% 1/16 iters=3, 1/8 iters=8 & 2.87 & 3.35 & 1.22 & 2.80 & 0.111 & 4.24 & 15.3 & 0.119 \\
% 1/16 iters=5, 1/8 iters=8 & 2.86 & 3.37 & 1.22 & 2.82 & 0.116 & 4.17 & 15.2 & 0.124 \\
% \hline
% 1/16 iters=1, 1/8 iters=4 & 3.14 & 3.63 & 1.31 & 2.76 & 0.082 & 5.58 & 17.9 & 0.088 \\
% 1/16 iters=1, 1/8 iters=6 & 3.02 & 3.50 & 1.26 & 2.69 & 0.094 & 4.66 & 15.9 & 0.101 \\
% \textbf{1/16 iters=1, 1/8 iters=8} & \textbf{2.97} & \textbf{3.44} & \textbf{1.24} & \textbf{2.67} & \textbf{0.106} & \textbf{4.33} & \textbf{15.3} & \textbf{0.114} \\
% 1/16 iters=1, 1/8 iters=10 & 2.95 & 3.41 & 1.23 & 2.68 & 0.119 & 4.17 & 15.0 & 0.127 \\
% \hline
% \end{tabular}
% \end{center}
% \caption{This table shows how different iterations affect both accuracy and inference time. The default configuration is 1 iteration of 1/16 refinement and 8 iterations of 1/8 refinement.}
% \label{table_3}
% \end{table*}

\begin{table*}[ht]
\begin{center}
\begin{tabular}{lcc|ccc|ccc}
\toprule
\multirow{2}{*}{\multirow{2}{*}{Refinement iterations}}& \multicolumn{2}{c}{Things} & \multicolumn{2}{c}{Sintel (train)} & \multirow{2}{*}{\multirow{2}{*}{\makecell{Jetson Orin\\Nano (s)}}} & \multicolumn{2}{c}{KITTI-15} & \multirow{2}{*}{\multirow{2}{*}{\makecell{Jetson Orin\\Nano (s)}}} \\
\cmidrule(r){2-3} \cmidrule(r){4-5} \cmidrule(r){7-8}
& train & val & clean & final & & EPE & F1 & \\
\midrule
\textbf{1/16 iters=1, 1/8 iters=8} & \textbf{2.97} & \textbf{3.44} & \textbf{1.24} & \textbf{2.67} & \textbf{0.106} & \textbf{4.33} & \textbf{15.3} & \textbf{0.114} \\
1/16 iters=3, 1/8 iters=8 & 2.87 & 3.35 & 1.22 & 2.80 & 0.111 & 4.24 & 15.3 & 0.119 \\
1/16 iters=5, 1/8 iters=8 & 2.86 & 3.37 & 1.22 & 2.82 & 0.116 & 4.17 & 15.2 & 0.124 \\
\midrule
1/16 iters=1, 1/8 iters=4 & 3.14 & 3.63 & 1.31 & 2.76 & 0.082 & 5.58 & 17.9 & 0.088 \\
1/16 iters=1, 1/8 iters=6 & 3.02 & 3.50 & 1.26 & 2.69 & 0.094 & 4.66 & 15.9 & 0.101 \\
\textbf{1/16 iters=1, 1/8 iters=8} & \textbf{2.97} & \textbf{3.44} & \textbf{1.24} & \textbf{2.67} & \textbf{0.106} & \textbf{4.33} & \textbf{15.3} & \textbf{0.114} \\
1/16 iters=1, 1/8 iters=10 & 2.95 & 3.41 & 1.23 & 2.68 & 0.119 & 4.17 & 15.0 & 0.127 \\
\bottomrule
\end{tabular}
\end{center}
\caption{This table shows how different iterations affect both accuracy and inference time. The default configuration is 1 iteration of 1/16 refinement and 8 iterations of 1/8 refinement.}
\vspace{-1em}
\label{table_3}
\end{table*}

\subsection{Ablation Study}
\label{sec: expt/ablation}

\textbf{Backbone Module:}
We found that using full scale features in the backbone actually does not help in estimating the 1/8th scale optical flow and overfits on the training dataset (Flying Things) leading to a slight drop in performance on both Sintel and KITTI datasets. Table \ref{ablation} shows this drop in performance when we add full scale features in the backbone. Moreover, using the YOLO v8 and the \predecessor\ backbone also lead to the same effect of overfitting on the training dataset and thus poor generalization on real-world data.

\textbf{Refine Module:}
We use 8 layers of CNN to output both refined optical flow and the hidden state in the refinement module. We experimented by reducing and adding 2 layers to observe the impact. The results show that reducing the number of layers slightly decreases accuracy, while adding layers does not improve accuracy. This indicates that eight layers provide a balanced configuration.

Our default feature dimensions are 128 for the 1/16 refinement and 96 for the 1/8 refinement. Reducing the feature dimensions by half (64 for the 1/16 refinement and 48 for the 1/8 refinement) results in a significant drop in accuracy.

In the refinement module, we use multiple layers of CNN to output the hidden state. We also experimented with replacing the first CNN layer with a ConvGRU to output the hidden state and using the remaining seven CNN layers to decode a refined flow. But including ConvGRU increases the computation burden by 20\% for the same number of iterations while being worse in performance. 

\textbf{Architecture:}
Cross-attention is used to exchange information between two input images globally. Removing it does not significantly affect accuracy on the KITTI dataset, but it causes a substantial drop in accuracy on the Sintel dataset.

Global matching provides the initial optical flow, which can handle large motions. Removing it and adding three more iterations of 1/16 refinement helps address large motion issues. The drop in accuracy indicates that global matching is working efficiently without requiring multiple refinements.

We only perform one iteration of 1/16 refinement. If we remove this and rely entirely on 1/8 refinement, overfitting occurs, as the training set accuracy remains unchanged while the accuracy drops significantly on all validation sets.

\textbf{Different Iterations:}
The default iteration count for 1/16 refinement is set to one, as additional iterations do not significantly improve accuracy. In contrast, 1/8 refinement benefits from more iterations. The default eight iterations already provide decent accuracy, but adding more iterations can further improve accuracy at the cost of increased inference time. See Fig. \ref{iters_plot} for more details.

\section{Conclusions and Future Work}

% In this paper, we proposed an efficient optical flow method, which accuracy is close to state-of-the-art while getting 10 times faster, enable real-time inference on edge computing device. However, we also recognized that memory consumption is heavy, which is caused by correlation computation. Various modules has addressed the problem \cite{morimitsu2024rapidflow}, \cite{morimitsu2024recurrent} which can be used in our architecture.

% Our method also contains too many parameters (9 million), primarily due to the simple backbone and simple RNN refinement module, which rely heavily on CNNs. This can potentially lead to overfitting with training data. Many efficient modules can be replaced to reduce the number of parameters. For example, MobileNets \cite{howard2017mobilenets}, \cite{sandler2018mobilenetv2}, \cite{howard2019searching} use depth-wise separable convolutions, while ShuffleNet \cite{zhang2018shufflenet} utilizes point-wise group convolution.

In this paper, we introduced \name\ , an efficient optical flow estimation framework that significantly enhances real-time performance without sacrificing accuracy. Our approach achieves competitive results while operating up to ten times faster than many existing methods, making it particularly well-suited for deployment on edge computing devices. The innovative design of our network—incorporating a lightweight backbone and an efficient iterative refinement module—demonstrates that high-quality optical flow estimation can be both efficient and practical for real-world applications. 
Comprehensive experiments validate the effectiveness of our method across various benchmarks, underscoring its potential for dynamic environments and resource-constrained settings. And more importantly, it paves the way for future research and practical applications. Looking ahead, we plan to further optimize memory efficiency and reduce parameter counts, thereby enhancing the model’s adaptability and scalability without compromising performance.

% Overall, our work represents a significant step toward bridging the gap between high-accuracy optical flow estimation and real-time inference, paving the way for future research and practical applications.

\bibliographystyle{IEEEtran}
\bibliography{IEEEabrv,mybib}

\end{document}